%% file: acl.tex
\pdfoutput=1
\documentclass[11pt]{article}

\usepackage{acl}

\usepackage{graphicx,times,latexsym,float,upquote,amsmath,booktabs,microtype,footnote}
\usepackage[linesnumbered,ruled,vlined]{algorithm2e}
\usepackage{array,etoolbox}
\preto\tabular{\setcounter{magicrownumbers}{0}}
\newcounter{magicrownumbers}

\usepackage[T1]{fontenc}
\usepackage[utf8]{inputenc}
\SetKwInput{KwInput}{Input}                
\SetKwInput{KwOutput}{Output}              
\SetKwInput{KwFunctions}{Functions}              

\let\origtau\tau 
\renewcommand{\tau}{\scalebox{1.44}{$\origtau$}}

\title{Automated Fact-Checking in Dialogue: Are Specialized Models Needed?}

\author{Eric Chamoun\textsuperscript{1}, Marzieh Saeidi\thanks{\textsuperscript{*} This work was completed while the author was at Meta AI.}
\textsuperscript{ 2} , Andreas Vlachos\textsuperscript{1}\\
         \textsuperscript{1}Department of Computer Science, University of Cambridge \\
         \textsuperscript{2}Synthesia, London \\
         \texttt{\{ec806,av308\}@cam.ac.uk}, \texttt{marzieh.saeidi@googlemail.com} }

\begin{document}
\makesavenoteenv{algorithm}
\maketitle

\begin{abstract}
Prior research has shown that typical fact-checking models for stand-alone claims struggle with claims made in dialogues. As a solution, fine-tuning these models on labelled dialogue data has been proposed.
However, creating separate models for each use case is impractical, and we show that fine-tuning models for dialogue results in poor performance on typical fact-checking. 
To overcome this challenge, we present techniques that allow us to use the same models for both dialogue and typical fact-checking. These mainly focus on retrieval adaptation and transforming conversational inputs so that they can be accurately predicted by models trained on stand-alone claims. We demonstrate that a typical fact-checking model incorporating these techniques is competitive with state-of-the-art models fine-tuned for dialogue, while maintaining its accuracy on stand-alone claims.
\end{abstract}
\begingroup
\let\clearpage\relax
\input{C1-Introduction/introduction.tex}
\input{C3-Methods/methods.tex}
\input{C5-Results/results.tex}
\input{conclusion.tex}
\endgroup
\bibliographystyle{acl_natbib}
\bibliography{acl}
\appendix
\renewcommand\thefigure{\thesection.\arabic{figure}} 
\renewcommand\thetable{\thesection.\arabic{table}}  

\include{Appendix/appendix.tex}
\end{document}

%% file: C1-Introduction/introduction.tex
\section{Introduction}
The need for fact-checking is ever-growing as the volume of false claims on social media platforms rises, inspiring researchers to develop automated tools to combat misinformation \citep{surveyZheng,surveyGuo}. 
Despite the application of automated fact-checking to various use cases, most studies still focus on stand-alone, well-formed claims similar to those found in formal sources like encyclopedias.
However, such claims are different from those found in conversations,
which often feature incomplete utterances that reference previously mentioned entities or even omit them \citep{cread,varshney}. 
Additionally, conversational utterances often include filler words and casual comments, which complicate the task. 

Recently, \citet{dialfact} presented DialFact, a dataset for automated fact-checking in dialogue. Their experiments showed that state-of-the-art models, trained on stand-alone well-formed claims, do not perform well on DialFact. 
To address this, they propose instead to fine-tune models on conversational claims within their dialogue contexts. 

In this paper, we first demonstrate that although these models obtain strong results on DialFact, they suffer from a significant decrease in accuracy on stand-alone fact-checking, a form of catastrophic forgetting, i.e.\ the tendency of a model to forget previously learned abilities after learning from new data \cite{catastrophic}. Furthermore, 
we argue that building a separate model for every real-world scenario is not a practical solution, since 
each 
model requires ongoing monitoring and maintenance for long-term reliability \citep{technicaldebt}.

For these reasons we introduce
methods for adapted evidence retrieval and input transformation without changing the fact-checking model. We first present a claim detection technique to tackle the low density of factual information in dialogue claims (Figure 1, in red). Additionally, we modify document retrieval to handle both the conversational context and the claim, but place more weight on the latter to reduce noisy retrieval results (Figure 1, in blue). Finally, we enhance sentence retrieval by considering not only the relevance of the retrieved sentence to the claim, but also that of the document it is sourced from (Figure 1, in green).
\begin{figure*}[!ht]
    \centering
    \includegraphics[width=16cm]{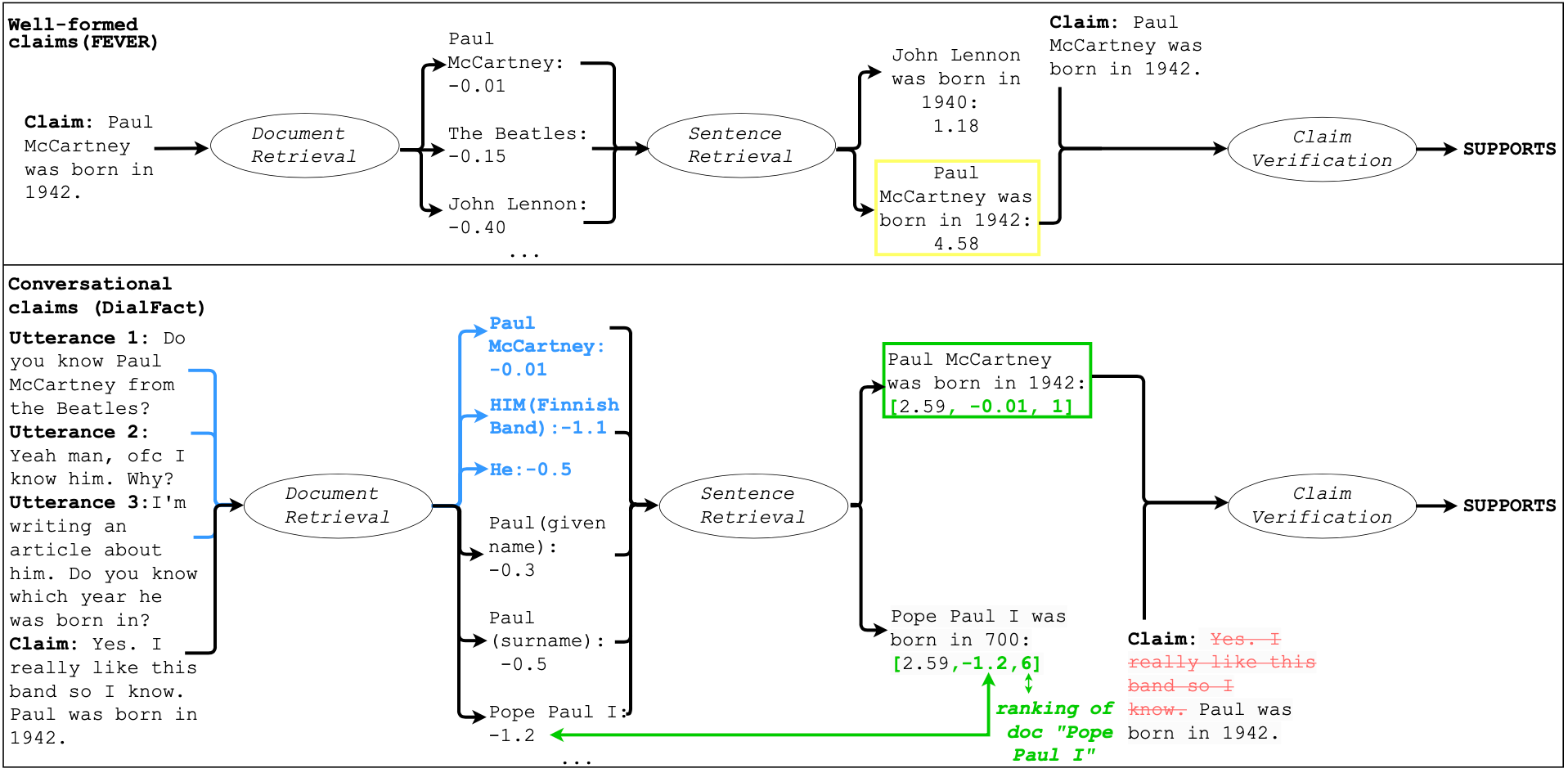}
    \caption{Overview of our approaches for typical and dialogue fact-checking. The proposed techniques are highlighted in blue for document retrieval, green for sentence retrieval and red for claim detection. }
    \label{fig:overview}
\end{figure*}

By incorporating the proposed techniques, a typical fact-checking model can match the performance of state-of-the-art models fine-tuned specifically for dialogue on DialFact, while  maintaining its accuracy on 
FEVER \cite{fever}, 
a benchmark for stand-alone well-formed claims
. In comparison, fine-tuning the same model for dialogue results in a minimum reduction of 12\% accuracy on FEVER due to catastrophic forgetting.

%% file: C3-Methods/methods.tex
\section{Fact-checking conversational claims}
Fact-checking systems typically consist of three components: a document retriever that returns the most relevant documents to a given claim from a textual source such as Wikipedia, a sentence retriever that selects the most relevant evidence sentence(s) from the retrieved documents, and a claim verification model that classifies the claim as \textsc{Supports}, \textsc{Refutes} or \textsc{Not Enough Info} (NEI) 
w.r.t.\ the evidence. We formulate fact-checking dialogue claims as the task of verifying the last utterance, referred to as the claim \textit{c}, of a multi-turn conversation $C = \{ U_1, ..., U_{n-1}, c\}$. 
This section presents techniques aimed at improving the dialogue fact-checking performance of a model trained on stand-alone well-formed claims, without requiring any adaptation of the model itself. 

\paragraph{Document retrieval}
In order to take into account the dialogue context of the claim,
we return the union of the top \textit{k} documents that are most similar to the claim, along with the single most relevant document to each utterance in the context:
\begin{equation}
    D_c = f(c,k)\text{ } \cup \sum_{i = 1}^{n-1} f(U_i,1) 
\end{equation},
where \textit{f} is a scoring function (not fine-tuned to dialogues) that takes as input a sentence \textit{s} and a number \textit{k}, and returns the top \textit{k} most relevant 
documents to \textit{s}.
The proposed method enables capturing the main entities of the conversation, despite the presence of coreference and ellipses. For example, in Figure 1, the name ``Paul McCartney'' is referred to as simply ``Paul'' in the conversational claim, making accurate document retrieval difficult.  Our approach tackles this challenge without retrieving a large number of irrelevant documents. Indeed, solely retrieving documents related to the claim would involve considering a broad range of entities containing ``Paul'' in their names. In contrast, the proposed method enables the retrieval of the precise entity under discussion within the conversation, eliminating the need to search through all possible ``Paul'' entities. Moreover, by focusing on the single most relevant document to each sentence within the context and simultaneously retrieving the top \textit{k} most relevant documents to the claim, we strike a balance. This approach ensures that the entities in the context are considered, but not to the extent that they overshadow the importance of the claim itself.

\paragraph{Sentence retrieval}
Just like document retrieval, the performance of sentence retrieval can be greatly impacted by the presence of coreference and ellipses in dialogues, as demonstrated in Figure 1. In the conversational example, the sentence retriever should 
assign the highest score to the Wikipedia sentence that contains information about the birth year of ``Paul''. However, in this case, ``Paul'' could equally refer to either ``Paul McCartney'' or ``Pope Paul I'', as documents with these titles were retrieved. As a result, the sentence retriever
is unable to distinguish the correct evidence and assigns equal scores to sentences from both documents.

To address this issue, we propose incorporating document relevance scores into sentence retrieval. This method capitalizes on the contextual information gathered during document retrieval, making sure it is utilized effectively in sentence retrieval. For instance, as shown in Figure 1, the information that ``Paul McCartney'' is the most relevant document to the conversation, while ``Pope Paul I'' is the sixth most relevant, increases the likelihood that the correct evidence is found in the former document. 

The proposed technique operates by combining information gathered during document and sentence retrieval as follows:
\begin{equation}
    score(s;c) = g(r_s; r_D; R_D)
\end{equation}
, where g is a parameterized function, s is an evidence sentence in a document \textit{D} $\in D_c$, $r_s$ is the similarity score of $s$ to the claim $c$, $r_D$ is the similarity score of \textit{D} to $c$, and $R_D$ is the ranking of \textit{D} among \textit{D$_c$}.  To train function \textit{g}, the document and sentence retrieval models are first applied to the DialFact training set examples to generate triples $(r_s; r_D; R_D)$ for each sentence \textit{s} with respect to a claim. Then, a logistic regression model is trained using these triples as inputs and Boolean values indicating whether each evidence sentence is a piece of evidence according to the gold standard.

\paragraph{Claim Detection} Typical fact-checking models are trained on claims that are single-factoid, self-contained sentences, such as those in FEVER~\cite{fever}. However, claims in dialogues often span multiple sentences, and may contain content that is not possible to check, such as ``Yes. I really like this band so I know.'' in Figure~1. This type of information can be challenging for these models to distinguish from the verifiable portion of the claim. 
To address this issue, we present a technique for identifying the factual information in dialogue claims. It operates by selecting the part of the utterance that has the highest semantic textual similarity to the retrieved evidence. The process begins by splitting the claim into sub-sentences. Next, we use a sentence encoder to generate encodings for each sub-sentence and each retrieved evidence sentence. Finally, the claim is replaced with the sub-sentence that has the highest cosine similarity with the retrieved evidence.

%% file: C5-Results/results.tex
\section{Results}
\paragraph{Implementation details} FEVER \cite{fever} is a benchmark dataset comprising well-formed claims derived from Wikipedia. Our approach builds on the state of the art. It employs GENRE\footnote{\url{https://github.com/facebookresearch/GENRE}} \cite{genre} as our scoring function  \textit{f} for document retrieval following \citet{genre-sota}. For sentence retrieval and claim verification, we leverage the Bigbird-based \cite{bigbird} and DeBERTa \cite{deberta} models\footnote{\url{https://github.com/dominiksinsaarland/document-level-FEVER}} respectively, from \citet{stammbach}. For the evidence enhancement ensemble, we train a logistic regression model \textit{g} using the scikit-learn library\footnote{\url{https://scikit-learn.org/stable/index.html}}. For claim detection, we leverage SRoBERTa\footnote{\url{https://github.com/UKPLab/sentence-transformers}} and Spacy's Sentencizer\footnote{\url{https://spacy.io/api/sentencizer/}} to perform sentence encoding and sentence splitting, respectively. Finally, 
we used SpanBERT \citep{spanbert} for coreference resolution, StyleFormer\footnote{
\url{https://github.com/PrithivirajDamodaran/Styleformer}} for style transfer, and the 
GPT-2-based model proposed by \citet{cread} for claim rewrite.

\paragraph{Document retrieval results}
\begin{table}[H]
\setcounter{table}{0}
\centering
\footnotesize
\begin{tabular}{@{}ccc@{}}
\toprule
                                                                 & \multicolumn{2}{c}{\textbf{Document Retrieval}}                             \\ \midrule
                                                                               & Recall & \begin{tabular}[c]{@{}c@{}}Recall\\ (No NEI)\end{tabular}  \\ \hline
  Claim-only                                                              & 56.85  & 60.02                                                     \\
  Resolved Claim-only                                                     & 67.0   & 72.0                                                      \\
  \begin{tabular}[c]{@{}c@{}}Concatenated \\ Claim + Context \cite{dialfact}\end{tabular} & 76.5   & 79.3                                                      \\
 Proposed Method                                                         & \textbf{81.90}  & \textbf{83.76}                                                     \\ \bottomrule
\end{tabular}
\caption{Document recall for claim-only and claim+context approaches using GENRE.}
\end{table} 
Table~1 presents a summary of the document retrieval results achieved on the DialFact test set. We conduct a comparative analysis of various methods: applying GENRE directly to the claim, applying GENRE to the claim after performing coreference resolution from the context using SpanBERT, employing the approach suggested by \citet{dialfact}, which entails concatenating the claim with the last two utterances of context and applying the scoring function \textit{f} to the result, and our proposed approach (Section 2). By looking at the results, it is clear that directly using the context substantially improves document recall. This result is expected, as the main entities of a conversation are often repeated numerous times in the context. Additionally, these methods do not depend on the coreference resolution accuracy. The scores also show that our proposed method improves upon that presented in \citet{dialfact} by more than 5 percentage points in terms of document recall when \textit{k = 10} (we select the top 10 documents using the claim). To ensure this performance increase is not merely due to an increase in the average number of retrieved documents, we additionally tested our method with \textit{k = 5}. The document recall, in this case, is equal to 80.07\% with an average number of retrieved documents of 7.79. This average includes 2.79 documents retrieved from the context, in addition to the top 5 documents most relevant to the claim. In contrast, other methods retrieve a minimum of 10 documents. The effectiveness of this approach compared to that presented in \citet{dialfact} can be explained by the higher emphasis on the claim. Our method focuses on retrieving the single most relevant document to each sentence within the context while simultaneously retrieving the top \textit{k} most relevant documents to the claim. In contrast, the approach in \citet{dialfact} directly applies the model to the concatenation of the claim and context, often resulting in the retrieval of noisy documents. 


\begin{table}
\centering
\footnotesize
\resizebox{\columnwidth}{!}{%
\begin{tabular}
{lccc}
\hline
 & \textbf{FEVER}  & \textbf{DialFact}                     \\ \hline
                     &  \multicolumn{2}{c}{\textbf{Accuracy}}            \\ \hline
FEVER & \textbf{79.80} & 50.75 (-12.88)    \\
\hspace{0.1in}+VitC   & 76.99 (-2.81) &  57.84 (-5.79)\\ \hline
\hspace{0.1in}+DialFact   & 67.03 (-12.77) &  61.08 (-2.55) \\
\hspace{0.1in}+Colloquial & 65.07 (-14.73) &  60.12 (-3.51)\\
\hspace{0.1in}+VitC+DialFact   & 64.88 (-14.92) &  61.99 (-1.64) \\
\hspace{0.1in}+VitC+Colloquial & 64.56 (-15.24) &  61.10 (-2.53) \\
\hline
\hspace{0.1in}\textit{+retrieval}  & \textbf{79.80} & 51.78 (-11.85) \\ 
\hspace{0.1in}\textit{+VitC+retrieval}   & 76.99 (-2.81) & 58.36 (-5.27) \\ 
\hspace{0.1in}\textit{+claimdet} & \textbf{79.80} &  53.30 (-10.33) \\
\hspace{0.1in}\textit{+VitC+claimdet} & 76.99 (-2.81) & 59.72 (-3.91)\\
\hspace{0.1in}\textit{+retrieval+claimdet} & \textbf{79.80} & 54.56 (-9.07)\\
\hspace{0.1in}\textit{+VitC+retrieval+claimdet} & 76.99 (-2.81) & 60.72 (-2.91) \\ \hline

\hspace{0.1in}+DialFact+retrieval & 67.03 (-12.77)  & 62.83 (-0.80) \\
\hspace{0.1in}+Colloquial+retrieval & 65.07 (-14.73) & 60.93 (-2.70) \\ 
\hspace{0.1in}+VitC+DialFact+retrieval & 64.88 (-14.92)  & \textbf{63.63} \\
\hspace{0.1in}+VitC+Colloquial+retrieval & 64.56 (-15.24) & 61.54 (-2.09) \\ 
\hline
AugWoW \cite{dialfact} & 60.98 (-18.82) & 51.60 (-12.03) \\ 
AugWoW+retrieval & 60.98 (-18.82) & 54.38 (-9.25) \\ 
\end{tabular}
}
\caption{Performance analysis on FEVER \textsc{Dev} and DialFact \textsc{Test} of typical fact-checking models combining our methods versus specialized models for dialogues. The  models proposed in this paper are in italic. Best performance per dataset is in bold.}
\label{tbl:results}
\end{table}

\paragraph{Final results} Table 2 summarizes the claim verification results on the test set. The first group consists of typical fact-checking models trained on FEVER (\textit{FEVER}), fine-tuned using VitaminC \citep[\textit{FEVER+VitC}]{dialfact}. VitaminC \citep{vitaminc} is a large-scale dataset containing examples that are \textit{contrastive}: evidence pairs are almost identical in language and content, except that one supports and the other refutes a claim. Training a fact-checking model on VitaminC has been shown to improve a classifier’s sensitivity to subtle changes in evidence. In our case, fine-tuning on VitaminC improves the DialFact accuracy by over 7\%, while only causing a decrease of less than 3\% on FEVER. 

In the second group, we fine-tune the typical models on additional dialogue data from DialFact and Colloquial \cite{colloquial}, as proposed by \citet{dialfact}. Specializing the models for dialogue leads to significant improvements in DialFact accuracy. However, this approach results in a substantial loss of up to 15\% in accuracy on FEVER due to catastrophic forgetting.

The third group uses the typical models with our proposed enhanced evidence selection and claim detection techniques. Applying these together leads to substantial performance improvements of up to 4\% on DialFact, while maintaining the accuracy on FEVER. This is reflected in the DialFact accuracy of \textit{FEVER+VitC+retrieval+claimdet}, which performs similarly to the top specialized models for dialogue fact-checking in group 2 while outperforming them by over 12\% on FEVER.

The next group demonstrates the advantages of incorporating our evidence enhancement technique to models fine-tuned for dialogue. 
Specifically, \textit{FEVER+VitC+DialFact+retrieval} achieves state-of-the-art performance on DialFact, outperforming the best previously published results \cite{dialfact} by 12\%.
\begin{figure}
    \centering
    \includegraphics[width = .9\linewidth ]{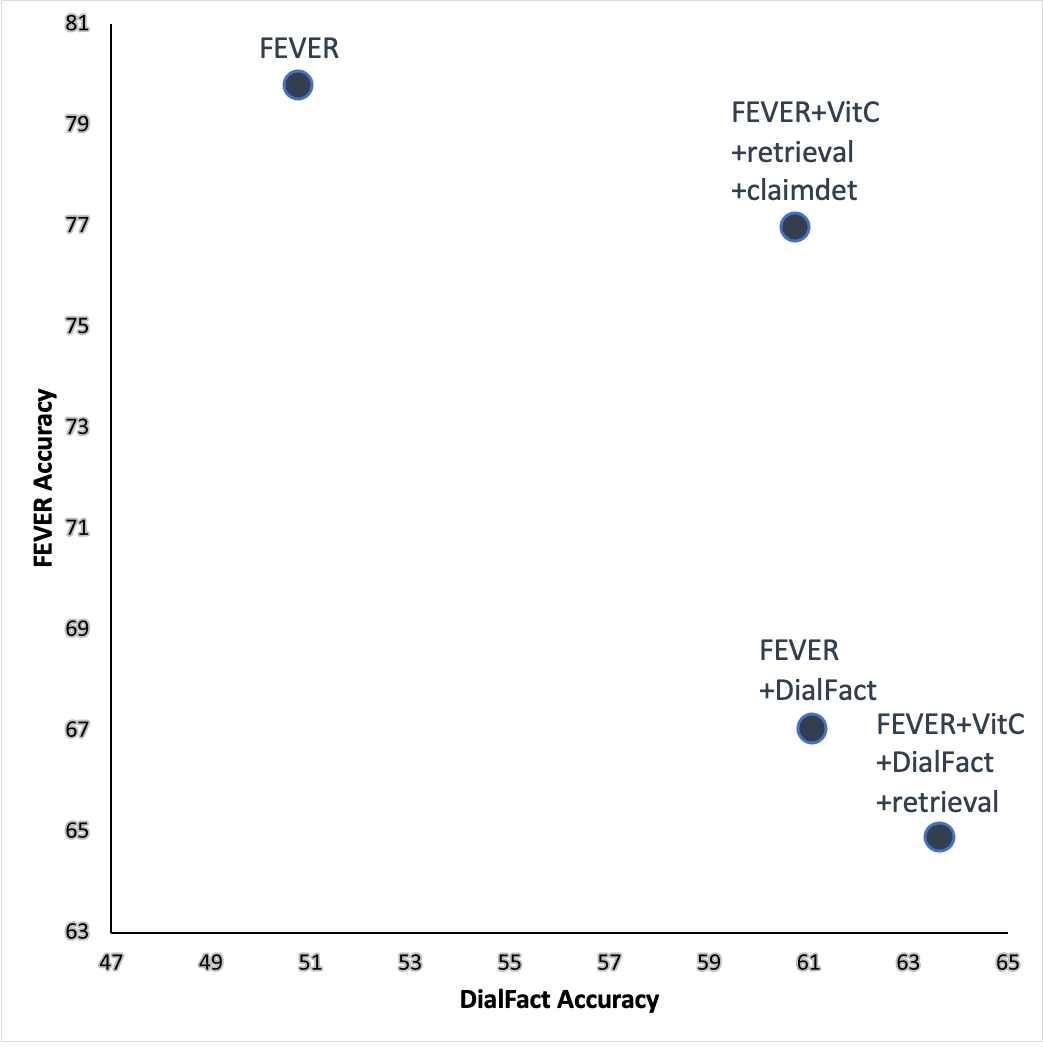}
    \caption{Tradeoff analysis between the accuracy scores on FEVER and DialFact for each model. }
    \label{fig:my_label}
\end{figure}

Finally, in the last group, we compare the DialFact accuracy of \textit{AugWow}, the top-performing model from \citet{dialfact}, when applied using evidence from the best specialized pipeline in their work, versus using our retrieved evidence. The results show an improvement of slightly less than 3\% when employing our retrieved evidence without using any specialized models for dialogue. 

Figure 2 illustrates the tradeoff between accuracy on dialogue fact-checking and catastrophic forgetting on FEVER. \textit{FEVER+VitC+retrieval+claimdet}, a model that combines our proposed methods, optimizes this balance better than the other approaches. It achieves near state-of-the-art results on both FEVER and DialFact, as seen by its placement in the upper right quadrant of the graph. In contrast, the state-of-the-art models for FEVER and DialFact are closer to one of the axes, reflecting their inferior performances on conversational claims and stand-alone formal claims respectively.

\section{Qualitative analysis}
We conducted a qualitative analysis to assess the advantages and limitations of different approaches.

We initially analyzed the performance improvements offered by our proposed techniques when incorporated to typical fact-checking models by focusing on cases where the central entities in the claim were referred to using pronouns. Our findings showed that most of the times, the document retrieval technique we proposed was still able to successfully identify the appropriate document by taking into account the context (Example~1 in Table~B.1).
In our study of claims with coreference, we encountered multiple situations where document retrieval was successful but returned multiple documents with  similar 
potential evidence sentences (e.g., birth years of ``Pope Paul I'' and ``Paul McCartney'', Figure~1). Our proposed sentence retrieval enhancement technique played a crucial role in these cases by using the context gathered during document retrieval to select the right evidence, resulting in more accurate predictions (Example~2 in Table~B.1). We also evaluated instances where the claim contained limited factual information. Frequently, our claim detection method effectively filtered out irrelevant sentences and allowed only the essential information to be processed by the model, leading to accurate predictions (Example 3 in Table B.1).

Additionally, we studied instances where specialized models for dialogue fact-checking outperformed the typical fact-checking model combining our techniques. We found that a common challenge in conversational claims that our approach does not address is indirect claims, such as those made in the form of a question. This limitation is due to the fact that typical fact-checking models are not trained to recognize indirect claims, and our proposed claim detection technique does not resolve this issue (Example 4 in Table B.1).

Finally, we examined cases where \textit{FEVER+VitC+retrieval+claimdet} outperforms specialized models for dialogue fact-checking. We discovered that in instances where a conversational claim could be easily transformed into a well-formed claim through methods such as claim detection, typical fact-checking models often outperformed those specifically designed for dialogue fact-checking. This is because the latter models experience catastrophic forgetting (Example 5 in Table B.1).

%% file: conclusion.tex
\section{Conclusion}
This paper highlights the significant catastrophic forgetting effects that result from adapting a typical fact-checking model for dialogue. We argue that using separate models for each task is not practical due to the ongoing maintenance cost attached to each. Instead, we propose techniques that  
allow us to use the same model for both use cases. These mainly focus on retrieval and input adaptation.
The model combining these techniques performs comparably to the top specialized models on DialFact while substantially outperforming them on FEVER. We discuss the limitations and societal impact of our approach in the Model Card (Figure A.1).

\section*{Limitations}
In this study, we present a model designed to autonomously verify claims extracted from dialogues. While our model demonstrates high accuracy on this specific task, it's important to acknowledge its limitations in real-world applications. Our testing benchmark, DialFact, consists of both human-generated and artificially constructed claims focused on a specific domain, utilizing Wikipedia as a knowledge base. However, this dataset's scope is confined to certain domains, which do not encompass all possible scenarios.

It's worth noting that despite the best efforts of \citet{dialfact} in ensuring the high quality of the annotation of DialFact, potential biases and inaccuracies could still exist as in all datasets. Additionally, our model's efficacy has been showcased solely on one dataset. As this task evolves and new datasets emerge, there's a necessity to evaluate our model's performance on diverse datasets to ensure its applicability across a range of scenarios.
\section*{Acknowledgements}
Eric Chamoun is supported by an EPSRC-funded studentship. Andreas Vlachos is supported by the ERC grant AVeriTeC (GA 865958).


%% file: Appendix/appendix.tex
\newpage
\section{Model Card}
\setcounter{figure}{0}
\begin{minipage}{1.0\textwidth}
  \strut\newline
  \centering
  \includegraphics[width=\textwidth]{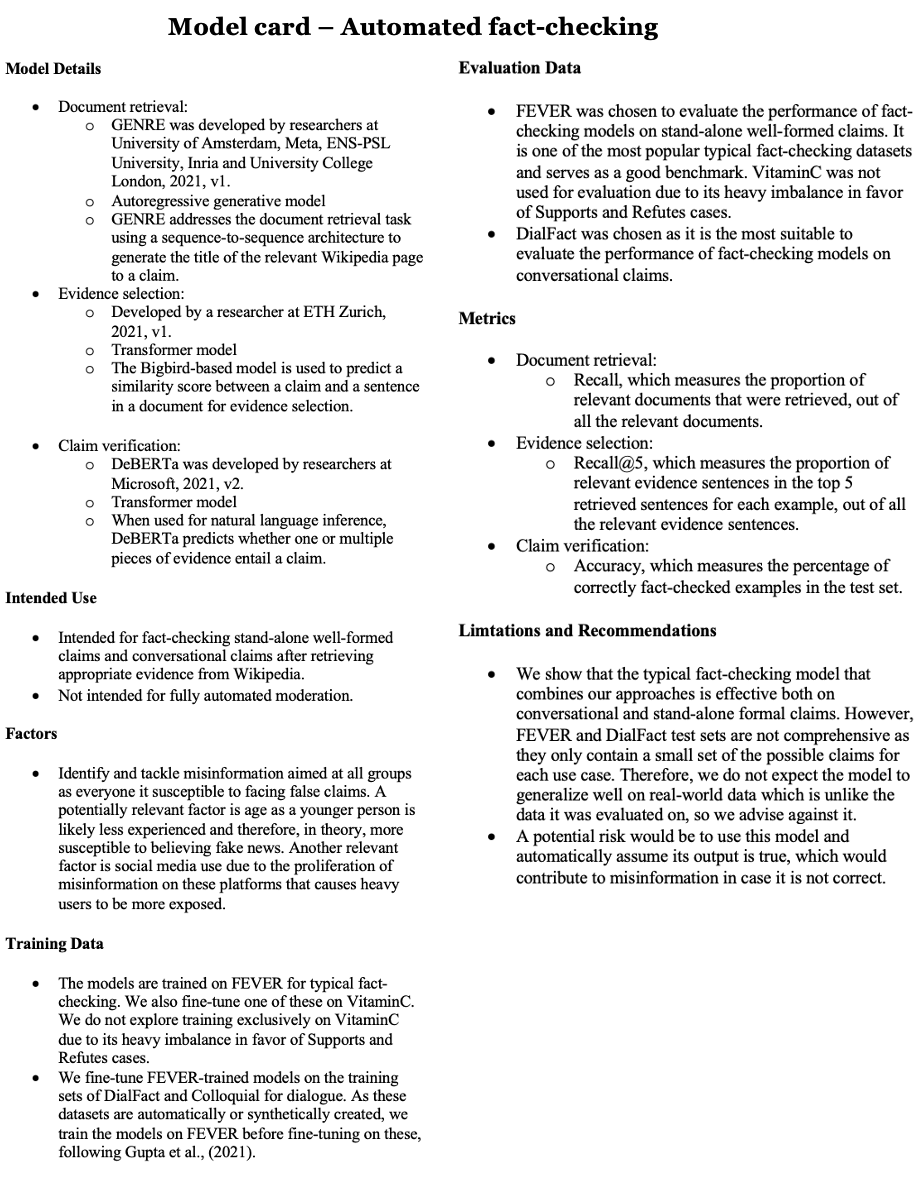}
  \captionof{figure}{Model card for the fact-checking model presented in this work.}\label{fig:figure1}
  \label{table:table A}
\end{minipage}
\newpage \clearpage
\section{Qualitative analysis examples}
\setcounter{table}{0}
\begin{minipage}{1.0\textwidth}
\vspace*{-1cm}
  \strut\newline
  \centering
  \includegraphics[width=\textwidth]{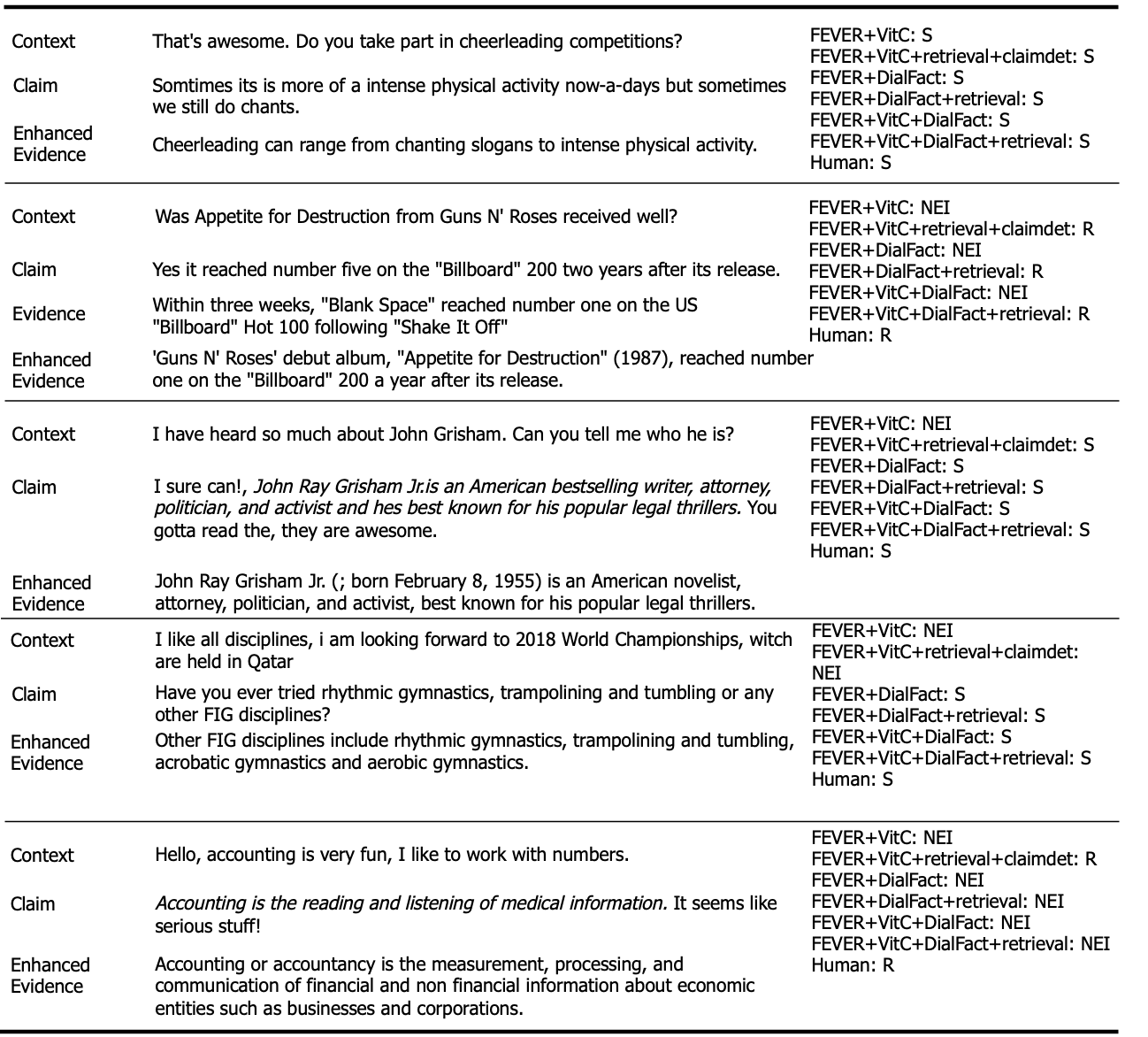}
  \captionof{table}{Sample \textsc{DialFact Test} instances highlighting the strengths and weaknesses of the top-performing models. S, R and NEI stand for \textsc{Supports}, \textsc{Refutes} and \textsc{Not Enough Info}, respectively. With the exception of Example 2, which is specifically chosen to demonstrate the advantages of our retrieval enhancement method, we have selected examples where the correct evidence is retrieved with or without the enhancement. To keep the context concise, we have only included the last turn of the conversation preceding the claim in these examples.}\label{fig:figure1}
  \label{table:table A}
\end{minipage}

The first example demonstrates the efficacy of our document retrieval technique. Despite the claim only referencing ``Cheerleading'' using a pronoun, the evidence required to verify it is effectively retrieved from the corresponding document, resulting in accurate predictions from all models.

The second example illustrates the advantages of the retrieval enhancement ensemble. Without using it, no evidence sentence related to ``Appetite for Destruction'' is found among the top 5 predictions. However, with the enhancement method, the crucial piece of evidence from the ``Guns N' Roses'' document is assigned the highest score. \\
\vspace*{17.75cm}\\
\noindent This is reflected in the claim verification results, as models that did not utilize the retrieval enhancement method produced incorrect predictions.

In the third example, all the models but \textit{FEVER+VitC} generate correct predictions. The specialized models are trained to eliminate extraneous information like ``I sure can'' and concentrate on the verifiable part of the claim (in italics). Additionally, \textit{FEVER+VitC+retrieval+claimdet} leverages claim detection before claim verification. This approach discards the first and last sentences, retaining only the factual information in the claim, which can be effectively processed by the typical fact-checking model.

The fourth example highlights a typical challenge in conversational claims that is not addressed by our proposed methods. The claim is made indirectly, in the form of a question, which typical fact-checking models are not trained to identify. This form of \textit{disguised} claim is not addressed either by our claim detection method, leading to \textsc{Not Enough Info} predictions from \textit{FEVER+VitC} and \textit{FEVER+VitC+retrieval-claimdet}. In contrast, models fine-tuned specifically for dialogue are able to effectively handle this challenge and generate accurate predictions.

Finally, the fifth example demonstrates a scenario where typical fact-checking models outperform models designed for dialogue fact-checking. All models except for \textit{FEVER+VitC+retrieval-claimdet} produce incorrect predictions. \textit{FEVER+VitC} fails to identify the claim's verifiable portion, while specialized models for dialogue fail to verify the claim with respect to the evidence. However, by applying the claim detection method, \textit{FEVER+VitC+retrieval-claimdet} is left with a well-formed formal claim that it verifies correctly. In cases where the conversational claim can be easily converted into a well-formed claim and does not present significant challenges in dialogue, typical fact-checking models can be more effective due to the catastrophic forgetting effects suffered by models fine-tuned for dialogue fact-checking.




\section{Claim-transformation techniques}
\citet{dialfact} state that the dialogue domain poses three main challenges for standard fact-checking models: the coreference and ellipsis phenomena, the low density of factual information in claims, and the colloquial language. In response, we proposed claim-transformation techniques to address these challenges directly.
Namely, we explored coreference resolution \cite{spanbert} and claim rewrite \cite{cread} to obtain self-contained claims that can be understood independent of previous dialogue context. These involve utilizing information from the previous turns to resolve coreference and ellipses. Additionally, we examined the benefits of applying style transfer to tackle the typical model's struggles with colloquial language. As these techniques were less effective than the techniques discussed in this paper, we include their results in the Appendix.
\subsection{Coreference resolution}
In DialFact, \citet{dialfact} choose to incorporate context by feeding models the concatenation of the claim and the last two utterances preceding it. However, this method requires not only specializing the model for the task but also adds significant noise. Nevertheless, context is crucial for claim understanding due to coreference and ellipses. Therefore, we propose directly addressing these issues by performing coreference resolution to obtain self-contained claims. 

We first concatenate the whole dialogue context with the claim. Subsequently, the coreference resolution model predicts coreference clusters in the resulting query. Each cluster consists of ((\textit{span start, span end), span tokens}) pairs, with the first pair being the referent and the remaining ones being its references. Subsequently, we use the span boundaries to replace each reference with its referent and obtain self-contained claims.

We present the results below.
\setcounter{table}{0}
\begin{table}[H]
\hspace{-0.2cm}
\resizebox{.5\textwidth}{!}{%
\begin{tabular}{cccccc}
\hline
\textbf{}       & \multicolumn{1}{c}{\begin{tabular}[c]{@{}c@{}}\textbf{Document}\\  \textbf{Retrieval}\end{tabular}}                              & \multicolumn{2}{c}{\textbf{Evidence Selection}}                                                                                                     & \multicolumn{2}{c}{\begin{tabular}[c]{@{}c@{}}\textbf{Claim Verification}\\  (Oracle Evidence)\end{tabular}} \\ \hline
                & Recall  & \begin{tabular}[c]{@{}c@{}}Verification \\ Accuracy \end{tabular} & Recall@5 & Accuracy           & Macro F1          \\ \hline
Untreated & 56.85                                                       & \textbf{54.19}                                                            & \textbf{44.06}                                                         & \textbf{58.73}              & \textbf{56.66}             \\
Resolved  & \textbf{67.0}                                                        & 53.86                                                            & 42.71                                                          & 58.60                   &    56.58              \\ \hline
\end{tabular}}
\smallskip
\caption{Impact of coreference resolution on each stage of the fact-checking process on DialFact \textsc{Dev}.}
\end{table}
Coreference resolution improves document retrieval. This result is unsurprising as replacing the mentions with their referents allows GENRE to identify the relevant entities and retrieve their documents. However, this technique harms the evidence sentence selection and claim verification performance. This negative impact can be explained by incorrect resolution cases where the reference is linked to the wrong referent. Indeed, evidence sentence selection is sensitive to a reference resolution mistake as it causes the model to select the most similar sentence in the wrong document. In claim verification, the model is slightly affected because a resolution error changes a claim's label with respect to the gold evidence. For each of these stages, the linking mistakes outweigh the advantages of using this method. In contrast, for document retrieval, the chances of accurate retrieval significantly increase if the resolution is correct.
\subsection{Claim rewrite}
Another approach we explore to obtain self-contained claims that can be understood independent of context is claim rewrite. 

We present the results of applying claim rewrite below.
\begin{table}[H]
\hspace{-0.4cm}
\resizebox{8.3cm}{!}{%
\begin{tabular}{cccccc}
\hline
\textbf{}       & \multicolumn{1}{c}{\begin{tabular}[c]{@{}c@{}}\textbf{Document}\\  \textbf{Retrieval}\end{tabular}}                              & \multicolumn{2}{c}{\textbf{Evidence Selection}}                                                                                                     & \multicolumn{2}{c}{\begin{tabular}[c]{@{}c@{}}\textbf{Claim Verification}\\  (Oracle Evidence)\end{tabular}} \\ \hline
                & Recall  & \begin{tabular}[c]{@{}c@{}}Verification \\ Accuracy \end{tabular} & Recall@5 & Accuracy           & Macro F1          \\ \hline
Untreated & 56.85                                                       & \textbf{54.19}                                                            & \textbf{44.06}                                                         & \textbf{58.73}              & \textbf{56.66}             \\
Resolved  & \textbf{59.56}                                                       & 53.67                                                            & 42.12                                                          & 58.53                   &   56.4               \\ \hline
\end{tabular}}
\smallskip
\caption{Impact of claim rewrite on each stage of the fact-checking process on DialFact \textsc{Dev}.}
\end{table}
Table C.2 shows a very similar pattern to the coreference resolution results. Indeed, claim rewrite improves document retrieval but harms evidence sentence selection and claim verification. However, a manual error analysis reveals numerous rewriting errors and an overall low resolution accuracy. This results in a poorer performance on all fact-checking subtasks than applying coreference resolution.
\subsection{Style transfer}
Spelling and punctuation mistakes, slang words and colloquialisms make it difficult for a model trained on formal claims to capture the intent of a colloquial claim. Another challenge is the presence of filler words, which significantly affects a retriever's ability to return the correct documents, as shown by \citet{colloquial}. Instead of retrieving relevant documents, the models return documents related to these filler words. In response, we explore style transfer in a bid to formalize the claims. The motivation behind this approach is that it would decrease the claims' wordiness and the presence of expressions that the model may find difficult to understand or recognize. Performing style transfer also expands abbreviations and corrects spelling mistakes and capitalization, which can be key to correct retrieval. 

We present the results of applying style transfer below.
\begin{table}[H]
\hspace{-0.2cm}
\resizebox{.5\textwidth}{!}{%
\begin{tabular}{cccccc}
\hline
\textbf{}       & \multicolumn{1}{c}{\begin{tabular}[c]{@{}c@{}}\textbf{Document}\\  \textbf{Retrieval}\end{tabular}}                              & \multicolumn{2}{c}{\textbf{Evidence Selection}}                                                                                                     & \multicolumn{2}{c}{\begin{tabular}[c]{@{}c@{}}\textbf{Claim Verification}\\  (Oracle Evidence)\end{tabular}} \\ \hline
                & Recall  & \begin{tabular}[c]{@{}c@{}}Verification \\ Accuracy \end{tabular} & Recall@5 & Accuracy           & Macro F1          \\ \hline
Untreated & \textbf{56.85}                                                       & \textbf{54.19}                                                            & \textbf{44.06}                                                         & \textbf{58.73}              & \textbf{56.66}             \\
Resolved  & 55.0                                                      & 53.52                                                            & 41.81                                                          & 55.76                   &   52.62               \\ \hline
\end{tabular}}
\smallskip
\caption{Impact of style transfer on each stage of the fact-checking process on DialFact \textsc{Dev}.}
\end{table}
Table C.3 shows that style transfer does not improve the model performance on any fact-checking component despite the high quality of the formalization. We identify two possible reasons to explain the performance dip caused by this technique. First and most importantly, the generation errors that cause a detail to be omitted in the formalized claim or an incorrect reformulation. Indeed, fact-checking is very sensitive to small changes in a claim. The generation of an equivalent claim needs to be semantically perfect to preserve all details. 
Consider Example C.1. The reformulation of this claim would score highly on most evaluation metrics for language generation. However, the subtle difference between the two claims that lies in the replacement of \textit{thinnest} with \textit{thin} changes the label of the claim with respect to evidence. Indeed, brown hair is thin compared to red hair but thick compared to fair hair in the gold evidence. In contrast, if we keep \textit{thinnest} then the claim is refuted by the evidence. 
\vspace{0.2cm}\\
\noindent\underline{Example C.1:}

\noindent\textbf{Original Claim:} Brown is the color of hair that is the thinnest.\\
\textbf{Formal Claim:} Brown is the hair color that is thin. \\
\textbf{Gold Evidence:} Its strands are thicker than those of fair hair but not as much as those of red hair. 
\vspace{0.2cm}\\
The second main reason follows from dataset construction. As the claims are created from Wikipedia passages containing the evidence sentences needed to verify the claims, these often use the same words or formulation as the evidences. This often facilitates the job of evidence sentence selection and claim verification. However, formalizing modifies these words, decreasing the similarity of a claim with its gold evidence. Consider Example C.2. Although the two sentences are semantically equivalent, the original one uses the same words as the evidence. 
\vspace{0.2cm}\\
\noindent\underline{\hypertarget{ex5}{Example C.2}:}
\vspace{0.2cm}\\
\noindent\textbf{Original Claim:} I wonder if this associates with the fact that darker hair is more common across the entire world. \\
\textbf{Formal Claim:} I am unsure if this is related to the widespread prevalence of darker hair in the world. \\
\textbf{Evidence:} Black hair is the darkest and most common of all human hair colors globally.